\documentclass{article}

\usepackage{arxiv}

\usepackage[T1]{fontenc}
\usepackage{sectsty}
\usepackage{subcaption}
\usepackage{titlesec}
\usepackage{graphicx}
\usepackage{todonotes}
\usepackage[backend=biber,sorting=nyt,bibencoding=latin1,style=numeric,maxnames=1,minnames=1]{biblatex}

\usepackage[stable]{footmisc}

\usepackage[utf8]{inputenc}
\usepackage[T1]{fontenc}

\usepackage{hyperref}
\usepackage{url}

\titlespacing*{\section}{0pt}{8pt}{8pt}

\hypersetup{colorlinks=true, urlcolor=blue, linkcolor=blue, filecolor=blue, citecolor=blue, breaklinks=true, 
}
\usepackage{footmisc}
\begin{document}

\title{ArCo: the Italian Cultural Heritage Knowledge Graph}

\author{Valentina Anita Carriero\\
Semantic Technologies Laboratory \\
Institute of Cognitive Sciences and Technologies \\
Italian National Research Council \\
Via San Martino della Battaglia 44, 00185 Rome, Italy \\ 
\texttt{valentina.carriero@istc.cnr.it} \And
Aldo Gangemi\\
Digital Humanities Advanced Research Centre \\
Department of Classical Philology and Italian Studies \\
University of Bologna \\
Via Zamboni 32, 00126 Bologna, Italy \\
\texttt{aldo.gangemi@unibo.it} \And
Maria Letizia Mancinelli \\
Central Institute for Cataloguing and Documentation \\
Ministry of Cultural Heritage and Activities \\
Via di San Michele 18, 00153 Roma \\
\texttt{marialetizia.mancinelli@beniculturali.it} \And 
Ludovica Marinucci \\
Semantic Technologies Laboratory \\
Institute of Cognitive Sciences and Technologies \\
Italian National Research Council \\
Via San Martino della Battaglia 44, 00185 Rome, Italy \\ 
\texttt{ludovica.marinucci@istc.cnr.it} \And 
Andrea Giovanni Nuzzolese \\
Semantic Technologies Laboratory \\
Institute of Cognitive Sciences and Technologies \\
Italian National Research Council \\
Via San Martino della Battaglia 44, 00185 Rome, Italy \\ 
\texttt{andreagiovanni.nuzzolese@.cnr.it} \And 
Valentina Presutti \\
Semantic Technologies Laboratory \\
Institute of Cognitive Sciences and Technologies \\
Italian National Research Council \\
Via San Martino della Battaglia 44, 00185 Rome, Italy \\ 
\texttt{valentina.presutti@.cnr.it} \And 
Chiara Veninata \\
Central Institute for Cataloguing and Documentation \\
Ministry of Cultural Heritage and Activities \\
Via di San Michele 18, 00153 Roma \\
\texttt{chiara.veninata@beniculturali.it}
}

\maketitle

\begin{abstract}
ArCo is the Italian Cultural Heritage  knowledge graph, consisting of a network of seven vocabularies and 169 million triples about 820 thousand cultural entities. It is distributed jointly with a SPARQL endpoint, a software for converting catalogue records to RDF, and a rich suite of documentation material (testing, evaluation, how-to, examples, etc.). 
ArCo is based on the official General Catalogue of the Italian Ministry of Cultural Heritage and Activities (MiBAC) - and its associated encoding regulations - which collects and validates the catalogue records of (ideally) all Italian Cultural Heritage properties (excluding libraries and archives), contributed by CH administrators from all over Italy.
We present its structure, design methods and tools, its growing community, and delineate its importance, quality, and impact.


\end{abstract}
\section{Bringing the Italian Cultural Heritage to LOD}
\label{sec:intro}
Cultural Heritage (CH) is the legacy of physical artifacts and intangible attributes of a group or society that is inherited from past generations. It carries aesthetical, social, historical, cognitive, as well as economic power. 
The availability of linked open data (LOD) about CH has already shown its potential in many application areas including tourism, teaching, management, etc. The higher the quality and richness of data and links, the higher the value that society, science and economy can gain from it. \\
As of July 2018, UNESCO has designated a total of 1092 World Heritage sites located in 167 different countries around the world, including cultural (845), natural (209), and hybrid (38) sites. According to UNESCO's list, Italy is the country with the highest number of world heritage sites (54)\footnote{42 additional sites are currently under review.}. UNESCO's list only indexes the tip of the iceberg of Italian CH, which is managed by the Italian Ministry of Cultural  Heritage and Activities (MiBAC). Within MiBAC, ICCD (Institute of the General Catalogue and Documentation) is in charge of maintaining a catalogue of (ideally) every item in the whole Italian CH (excluding libraries and archives), as well as to define standards for encoding catalogue records describing them, and to collect these records from the diverse institutions that administer cultural properties throughout the Italian territory. To date, ICCD has assigned more than 15M unique catalogue numbers to its contributors (cf. Section \ref{sec:sigec} for further details), and has collected and stored $\sim$2.5M records, $\sim$0.8M of which are available for consultation on its official website.
This growing, standardised, curated catalogue is the heart of Italian CH data, and a potential \emph{hub} for a highly reliable and rich \emph{knowledge graph} of Italian CH, and beyond.

This paper describes ArCo, a new resource that realises this potential. ArCo is openly released with a \href{https://creativecommons.org/licenses/by-sa/4.0/}{CC-BY-SA 4.0 licence} both on \href{https://github.com/ICCD-MiBACT/ArCo/tree/master/ArCo-release}{GitHub}\footnote{The links are hidden by blue clickable words. You can find the complete list of URLs included in this paper at: \url{https://bit.ly/2VpR5cQ}}
and on the official \href{http://dati.beniculturali.it/}{MiBAC website}. It can be downloaded as a docker container and locally installed, or accessed \href{http://w3id.org/arco/}{online}.
ArCo includes:
\begin{itemize}
  \item a knowledge graph\footnote{There is no consensus on a definition for knowledge graph~\cite{bonatti_et_al:DR:2019:10328}, in this context we refer to linked open data including both OWL and RDF entities, and both schema axioms and factual data.} consisting of:
    \begin{itemize}
      \item a network of ontologies (and ontology design patterns), modeling the CH domain (with focus on cultural properties) at a fine grained level of detail (cf. Section \ref{sec:ontology});
      \item a LOD dataset counting $\sim$169M triples, which describe $\sim$0.8M cultural properties and their catalogue records;
    \end{itemize}
  \item a software for automatically converting catalogue records compliant to ICCD regulations to ArCo-compliant LOD, which enables automatic and frequent updates, and facilitates reuse;
  \item a detailed documentation reporting: (i) the ontological requirements, expressed in the form of user stories as well as competency questions (CQs), (ii) the resulting ontological models with diagrams and examples of usage;
  \item a set of running examples that potential consumers can use as training material. They consist of natural language CQs and their corresponding SPARQL queries, which can be directly tested against ArCo's SPARQL endpoint;
  \item a test suite, implemented as OWL files and SPARQL queries, used for validating ArCo knowledge graph (KG). It provides a real-case implementation of an ontology testing methodology, useful to both students, teachers, researchers, and practitioners.
  \item a SPARQL endpoint to explore the resource, run tests, etc.
\end{itemize}
It is worth remarking that ArCo data are of highly reliable provenance (cf. Section \ref{sec:sigec}). Its ontology network shows high quality, as resulting both as a project of eXtreme Design (XD), an established methodology~\cite{Blomqvist2010} based on the reuse of ontology design patterns (ODP) (cf. Section \ref{sec:method}), and emerging from an ex-post evaluation described in Subsection \ref{sec:eval}.
ArCo ontologies (cf. Section \ref{sec:ontology}) address, and are evaluated against, requirements elicited from both the data provider (ICCD), and a community of independent consumer representatives, including private and public organisations working with CH open data. These requirements have raised the need of new ontology models, which have been developed while reusing or aligning to relevant CH ontologies such as CIDOC-CRM~\cite{DBLP:journals/aim/Doerr03} and EDM~\cite{DBLP:journals/semweb/IsaacH13}. ArCo data (cf. Subsection \ref{sec:dataset}) links to $\sim$18.7K entities belonging to other LOD datasets, e.g. DBpedia, Wikidata, Geonames. \\
ArCo is reused in a separate project that involves ICCD and \href{https://artsandculture.google.com/}{Google Arts \& Culture},
focused on photographic cultural properties. Indeed, there is evidence of a growing community reusing and interested in ArCo as discussed in Section \ref{sec:impact}. ArCo is published by following FAIR principles (cf. Section \ref{sec:availability}). It is an evolving project, therefore there are many aspects that can and will be improved: they are briefly discussed in Section \ref{sec:conclusion}.
%
\section{The official catalogue of Italian Cultural Heritage}
\label{sec:sigec}

ArCo data derive from the \href{http://www.catalogo.beniculturali.it/sigecSSU_FE/Home.action?timestamp=1521647516354}{General Catalogue of Italian Cultural Heritage} (GC), the official institutional database of Italian CH, maintained and published by ICCD. GC currently contains 2.735.343 catalogue records, 781.902 of which are publicly consultable through the ICCD website. The remaining records may refer to private properties, or to properties being at risk (e.g. items in churches that are not guarded), or to properties that need to be scientifically assessed by accounted institutions, etc. GC is the result of a {\it collaborative effort} involving many and diverse contributors (currently 487) {\it formally} authorised by ICCD. These are national or regional, public or private, institutional organisations that administer cultural properties all over the Italian territory. They submit their catalogue records through a collaborative platform named \href{http://www.iccd.beniculturali.it/it/sigec-web}{SIGECweb}. Submissions undergo an automatic validation phase, aimed to check compliance with cataloguing standards
provided by ICCD for all kinds of {\it cultural properties}. A second scientific validation is performed by appointed experts. The authorisation and validation processes guarantee high standard for quality and provenance reliability of GC data as a source for ArCo. \\
In addition to GC data, ArCo's input includes requirements deriving from consumers' elicited use cases (cf. Section \ref{sec:method}), and \href{http://www.iccd.beniculturali.it/it/normative}{ICCD cataloguing standards} that define many types of {\it cultural properties}, precisely: archaeological, architectural and landscape, demo-ethno-anthropological, photographic, musical, natural, numismatic, scientific and technological, historical and artistic properties. 

\begin{figure}[ht!] \centering
\includegraphics[width=\textwidth,scale=0.40]{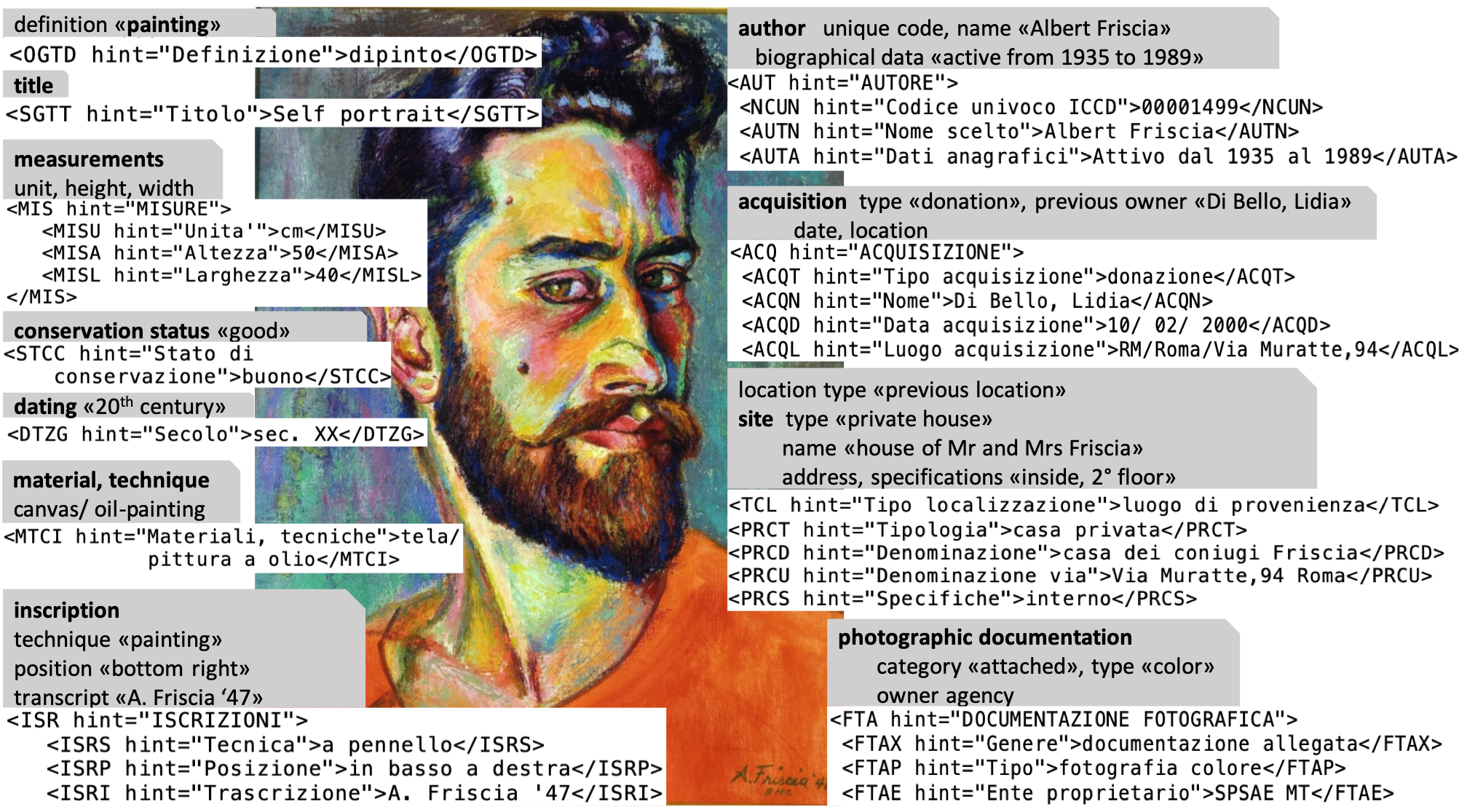}
	\caption{An example of XML data from ICCD Catalogue. Each snippet is translated to English.}
	\label{img:record}
\end{figure}

Figure \ref{img:record} 
depicts a painting by \href{https://en.wikipedia.org/wiki/Albert_J._Friscia}{Albert Friscia} with some excerpts from its XML catalogue record, from GC. 
%
The first issue with XML data is that all information in the records is expressed as strings. In order to build an RDF KG, good practice suggests to produce individuals for every element (or set of elements), whose value (or set of values) refers to anything that may participate as a subject in a triple, or can be linked to, or from, external resources. For example, we want to create an individual for Albert Friscia (the author), as well as an individual for the technical status of this painting. As part of the modeling process (cf. Section \ref{sec:ontology}), we define ArCo classes by abstracting from sets of fields and we define rules for creating URIs for their individuals 
(cf. Section \ref{sec:method}).

\section{Using eXtreme Design for developing ArCo\\ Knowledge Graph}
\label{sec:method}
ArCo knowledge graph (KG) is developed by following eXtreme Design (XD), focused on ontology design patterns (ODPs) reuse~\cite{Blomqvist2010}. 
XD is iterative and incremental, implementing a feedback loop cycle by involving different actors: (i) a \emph{design team}, in charge of selecting and implementing suitable ODPs as well as to perform alignments and linking; (ii) a \emph{testing team},  disjoint from the design team, which takes care of testing the ontology; (iii) a \emph{customer team}, who elicits the requirements that are translated by the design team and testing team into ontological commitments (i.e. competency questions and other constraints) that guide the ontology development.  

\begin{figure}[ht!] \centering
\includegraphics[scale=0.37]{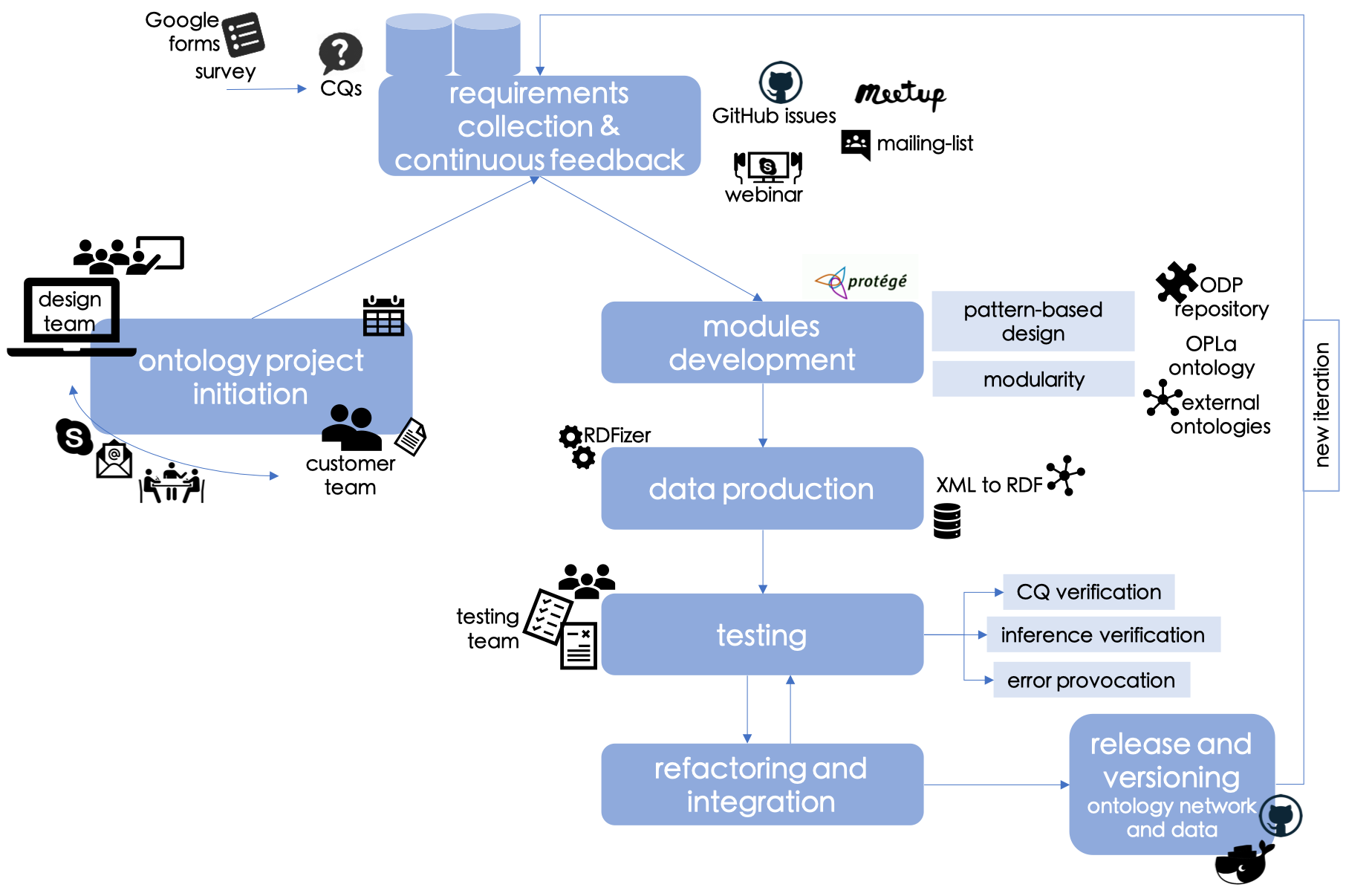}
	\caption{The XD methodology as implemented for the ArCo knowledge graph.}
	\label{img:methodology}
\end{figure}

Figure \ref{img:methodology} depicts as XD applied to ArCo, jointly with the tools used in the process, e.g. GitHub, Protégé, etc. The remainder of this section provides a detailed explanation of how each phase is implemented.

\noindent{\bf Ontology project initiation.}
The design team and the customer team (initially composed of experts from ICCD) have shared their knowledge about the domain, the data and the method, and have agreed on a release plan and on communication means.

\noindent{\bf Requirements collection and continuous feedback.}
ArCo's requirements are collected in the form of small stories (according to XD). They are then reformulated as \href{https://github.com/ICCD-MiBACT/ArCo/blob/master/ArCo-release/test/CQ/CQs-SPARQLqueries.txt}{Competency Questions} (CQs, cf.~\cite{Blomqvist2010}), and used for ODP selection by the design team, as well as in the testing phase, by the testing team (more in the remainder of this section). Stories are submitted by the customer team to a \href{https://goo.gl/forms/zCixt3B1ABYbj9JS2}{Google Form}. In order to capture a wider perspective on requirements than the institutional and regulatory ones, we extended the customer team by involving a number of representative stakeholders such as private companies and public administrations working with CH data, in addition to the data owner (ICCD). \\
Improvement proposals and bugs are collected as \href{https://github.com/ICCD-MiBACT/ArCo/issues}{issues through GitHub}. As ArCo is published with incremental releases (cf. Releases and versioning), the customer and the testing teams can contribute continuous and updated feedback, which allows the design team to early detect new emerging requirements and errors, and schedule them for next releases. 
A growing community, involving interested stakeholders and consumers, interacts {\it via} a dedicated \href{https://groups.google.com/forum/#!forum/arco-project}{mailing-list},
as well as by participating at meetups and webinars. \\
%
\noindent{\bf Module design.}
Ontology design patterns~\cite{DBLP:books/ios/HGJKP2016} play a central role in ArCo's design as recommended by XD. We adopt both {\it direct reuse} (i.e. reusing patterns from other ontologies by embedding their implementations in the local ontology) and {\it indirect reuse} (i.e. reusing patterns from other ontologies as templates, and adding alignment axioms to them). We reuse patterns from online repositories, e.g. \href{http://www.ontologydesignpatterns.org}{ODP portal}, and from existing ontologies, e.g. CIDOC-CRM. For details about indirect and direct reuse, the reader is invited to consult ~\cite{DBLP:conf/er/PresuttiLNGPA16}. In some cases we have developed new ODPs, as in the case of modeling recurrent events.
All (re)used ODPs in ArCo are annotated with OPLa ontology~\cite{DBLP:conf/semweb/HitzlerGJKP17}, which facilitates future reuse of ArCo as well as matching to other resources.
XD encourages and supports a modular design, where each ontology module addresses a subset of requirements and covers a coherent sub-area of the domain. Therefore, ArCo ontology network consists of seven modules, each with its own namespace.

\noindent{\bf Data production.}
ArCo RDF data are produced with \href{https://github.com/ICCD-MiBACT/ArCo/tree/master/ArCo-release/rdfizer}{RDFizer}. Its core component is {\it XML2RDF Converter}, which takes two inputs: an XML file compliant with \href{http://www.iccd.beniculturali.it/it/normative}{ICCD cataloguing standards}, 
and XSLT stylesheets specifying how to map XML tags to RDF. Its output is an RDF dump used to feed a triplestore.\\
\noindent{\bf URI production.} Let us consider generating the URI for the author \emph{Albert Friscia}, referring to Figure \ref{img:record}.
ArCo base URI for individuals is \texttt{https://w3id\-.org\-/ar\-co\-/re\-sour\-ce/} with prefix  \texttt{data:}. Every individual's ID is preceded by the name of its type, e.g. \texttt{Agent}. For each type, we manually identify a set of elements that constitute a possible key (e.g. \texttt{AUTN}, the author's name). We remove punctuation from the values of these elements (which are strings), convert them in lower case, concatenate them and sort them in alphabetical order, e.g. \texttt{albert-friscia}. We compute an MD5 checksum on the resulting string, which is used as the URI's ID e.g, \href{https://w3id.org/arco/resource/Agent/dcd4ca7b54dd3d7dac083dd4c54a9eef}{data:Agent/dcd4ca7b54dd3d7dac083dd4c54a9eef}. Some types have a unique identifier, e.g. cultural properties. In those cases we directly use it as the URI's ID. In order to minimise duplication, we then perform an entity linking step on the resulting individuals by using LIMES, with the same approach used for linking to external datasets (cf. Subsection \ref{sec:dataset}). 
\noindent{\bf Testing.}
To detect any incoherence in the ontologies, we regularly run a reasoner,  \href{http://www.hermit-reasoner.com/}{HermiT}, during the modeling phase. Then, to evaluate the appropriateness of the ontologies against requirements, we follow the methodology described in \cite{Blomqvist2012}, which focuses on testing an ontology against its requirements, intended as the ontological commitment expressed by means of CQs and domain constraints. All testing activities and resulting data (OWL files complying to the ontology described in \cite{Blomqvist2012}) are documented in a specific section of \href{https://github.com/ICCD-MiBACT/ArCo/tree/master/ArCo-release/test}{ArCo's GitHub repository}. The testing activity is iterative, and goes in parallel with the modeling activity (XD is test-driven). The testing team performs iterative testing, applying three approaches: \textit{CQ verification, inference verification, error provocation}. 
{\it CQ verification} consists in testing whether the ontology vocabulary allows to convert a CQ, e.g. ``When was a cultural property created, and what is the source of its dating?'' to a SPARQL query.
CQ verification allows to detect any missing concept or gap in the vocabulary (e.g. whether the class for representing the source of a {\it dating} has been modeled). 
{\it Inference verification} focuses on checking expected inferences. 
For example, if a \texttt{\href{https://w3id.org/arco/ontology/arco}{:}ComplexCulturalProperty} is defined as a  \texttt{:Cultural\allowbreak{}Property} that has one or more \texttt{:CulturalPropertyComponent}s, an axiom stating that a \texttt{:Cultural\allowbreak{}Property} has a \texttt{:CulturalPropertyComponent} would suffice to infer that the property is complex, even if it is not explicitly asserted.
But if the reasoner does not infer this information, this means that the appropriate axiom (in this case an equivalence axiom) is missing from the ontology.
{\it Error provocation} is intended to ``stress'' the knowledge graph by injecting inconsistent data. 
E.g. when characterising a \texttt{:Cultural\allowbreak{}Property}, an individual belonging to both \texttt{\href{https://w3id.org/arco/ontology/catalogue}{a-cd:}AuthorshipAttribution} and \texttt{a-cd:Dating} classes,
which are supposed to be disjoint, should result as inconsistent. If the reasoner does not detect the injected error, it means that the appropriate (disjointness) axiom is missing. 

\noindent{\bf Refactoring and integration.}
Problems spotted during the testing phase are passed back to the design team as issues. The design team refactors the modules and updates the ontology after performing a consistency checking. The result of this step is validated again by the testing team before including the model in the next release.


\noindent{\bf Releases and versioning.}
Incremental versions of ArCo KG are periodically and openly released. Every release has a version number, and each ontology module is marked with its own {\it version number} and {\it status}, the latter being one of: (i) {\it alpha} if the module has partly passed internal testing, (ii) {\it beta} if internal testing has been thoroughly performed, and testing based on external feedback is ongoing and partly fulfilled, (iii) {\it stable} when both internal and external testing have been thoroughly done. Releases are published as Docker containers on \href{https://github.com/ICCD-MiBACT/ArCo/tree/master/ArCo-release}{GitHub} and \href{https://w3id.org/arco}{online}.

\section{ArCo Knowledge Graph}
\label{sec:ontology}

ArCo's main component is a knowledge graph (KG), intended as the union of the ontology network and LOD data. Nevertheless, ArCo KG is released as part of a package including accompanying material (documentation, software, online services) that support its consumption, understanding and reuse. In this Section, we firstly detail what an ArCo release contains, and then we provide details about ArCo KG.

\subsection{How to use ArCo}
\label{sec:howto}
Each release of ArCo consists of a \emph{docker} container available on \href{https://github.com/ICCD-MiBACT/ArCo/tree/master/ArCo-release}{GitHub}, and its running instance \href{https://w3id.org/arco}{online} - both English and Italian versions. 
Each release contains:
\begin{itemize}
\item {\bf User guides} for supporting users in understanding the content of each release, with \href{http://www.essepuntato.it/graffoo}{Graffoo} diagrams and narrative explanations of every ontology module.   
\item {\bf Ontologies}, including their source code and a human-readable HTML documentation created with \href{http://www.essepuntato.it/lode}{LODE}.
\item {\bf A SPARQL endpoint} storing ArCo KG. The SPARQL endpoint also includes LOD data about \href{http://www.beniculturali.it/mibac/opencms/MiBAC/sito-MiBAC/MenuPrincipale/LuoghiDellaCultura/Ricerca/index.html}{cultural institutes or sites} and \href{http://www.beniculturali.it/mibac/opencms/MiBAC/sito-MiBAC/MenuPrincipale/EventiCulturali/Ricerca/index.html}{cultural events}, extracted from ``DB Unico 2.0''. We use \href{https://lodview.it/}{Lodview} as RDF viewer.
\item {\bf Examples of CQs} (cf. Section \ref{sec:method}) that ArCo KG can answer, with their corresponding SPARQL queries. This helps users to have a quick understanding of what is in ArCo ontologies and data, and how to use it.
\item {\bf A RDFizer tool} converting XML data represented according to \href{http://www.iccd.beniculturali.it/it/normative}{ICCD cataloguing standards} to RDF complying to ArCo ontologies.
\end{itemize}

\subsection{ArCo Ontology Network}
\label{sec:network}
\begin{figure}[ht!] \centering
\includegraphics[scale=0.45]{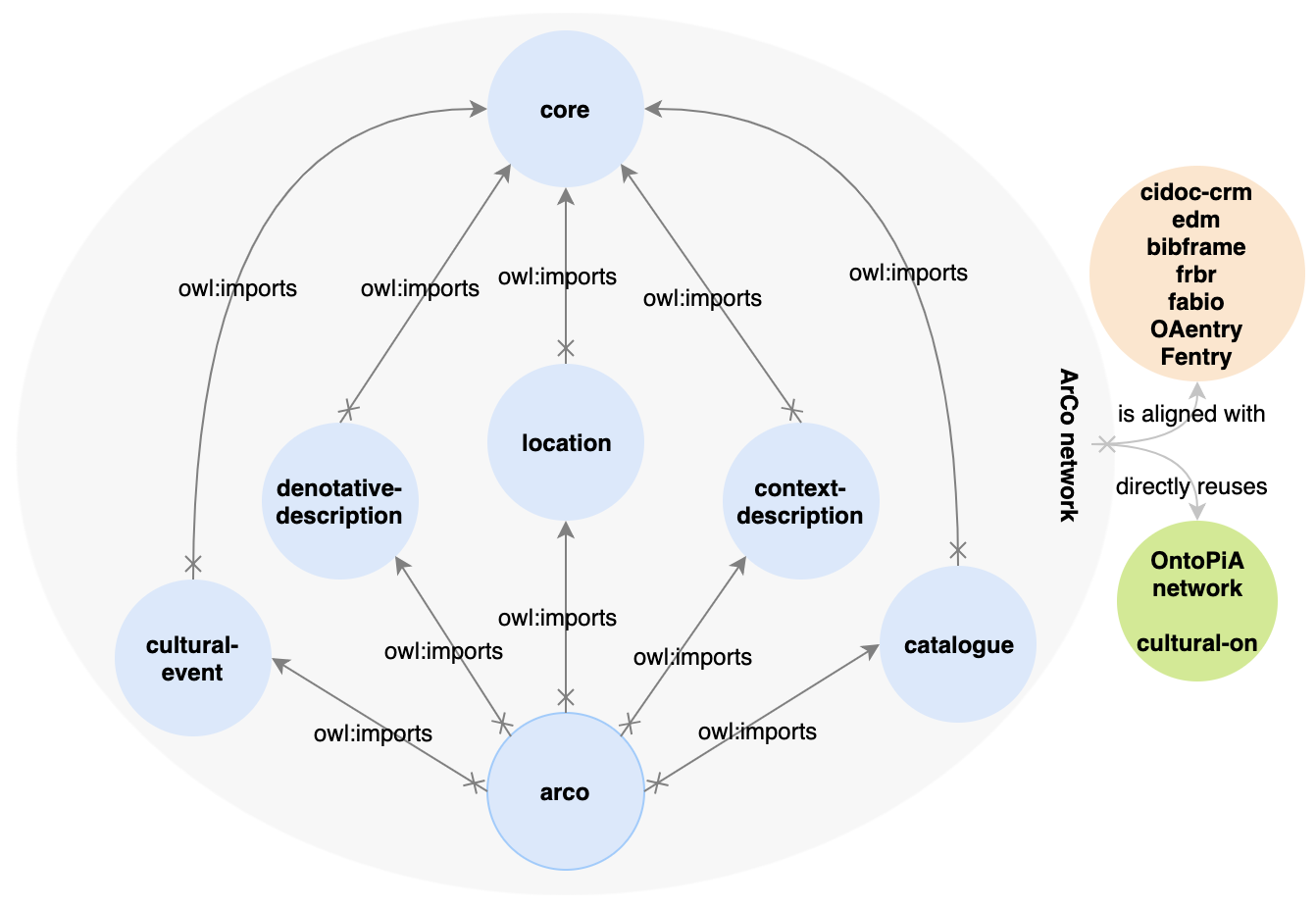}
	\caption{ArCo ontology network. Blue circles depict ArCo modules, the green circle indicates directly reused ontologies (they are embedded into ArCo), the orange circle indicates indirectly reused ontologies (some of their patterns are reused as templates, and alignment axioms are provided).}
	\label{img:arco-network}
\end{figure}

ArCo ontology network consists of 7 ontology modules connected by \texttt{owl:imports} axioms (cf. Figure \ref{img:arco-network}).
Two modules -- {\bf arco} and {\bf core} -- include top-level concepts and cross-module generic relations respectively. The {\bf catalogue} module is dedicated to catalogue records: not only do ArCo ontologies represent cultural properties, but also their ICCD catalogue records, in order to preserve the provenance and dynamics of the data. The remaining four modules ({\bf cultural-event, denotative-description, location, context-description}) focus on cultural properties and their features.

The network base namespace is \texttt{https://w3id.org/arco/ontology/}, which is shared by all modules (for each module we indicate its specific namespace). 

ArCo ontologies
define 327 classes, 379 object properties, 154 datatype properties, 176 named individuals (at the schema level). ArCo \emph{directly} reuses \href{http://dati.beniculturali.it/cis/}{Cultural-ON}
and \href{https://github.com/italia/daf-ontologie-vocabolari-controllati}{OntoPiA} (the ontology network for Italian Public Administrations),
while it \emph{indirectly} reuses \href{http://cidoc-crm.org/}{CIDOC-CRM},
\href{https://pro.europeana.eu/page/edm-documentation}{EDM},
\href{http://id.loc.gov/ontologies/bibframe/}{BIBFRAME},
\href{http://vocab.org/frbr/core}{FRBR},
\href{
http://purl.org/spar/fabio}{FaBiO},
\href{http://www.essepuntato.it/2014/03/fentry}{FEntry},
\href{http://purl.org/emmedi/oaentry}{OAEntry}.

\begin{figure}[ht!]
\begin{subfigure}{\textwidth}
  \centering
  \includegraphics[width=\textwidth]{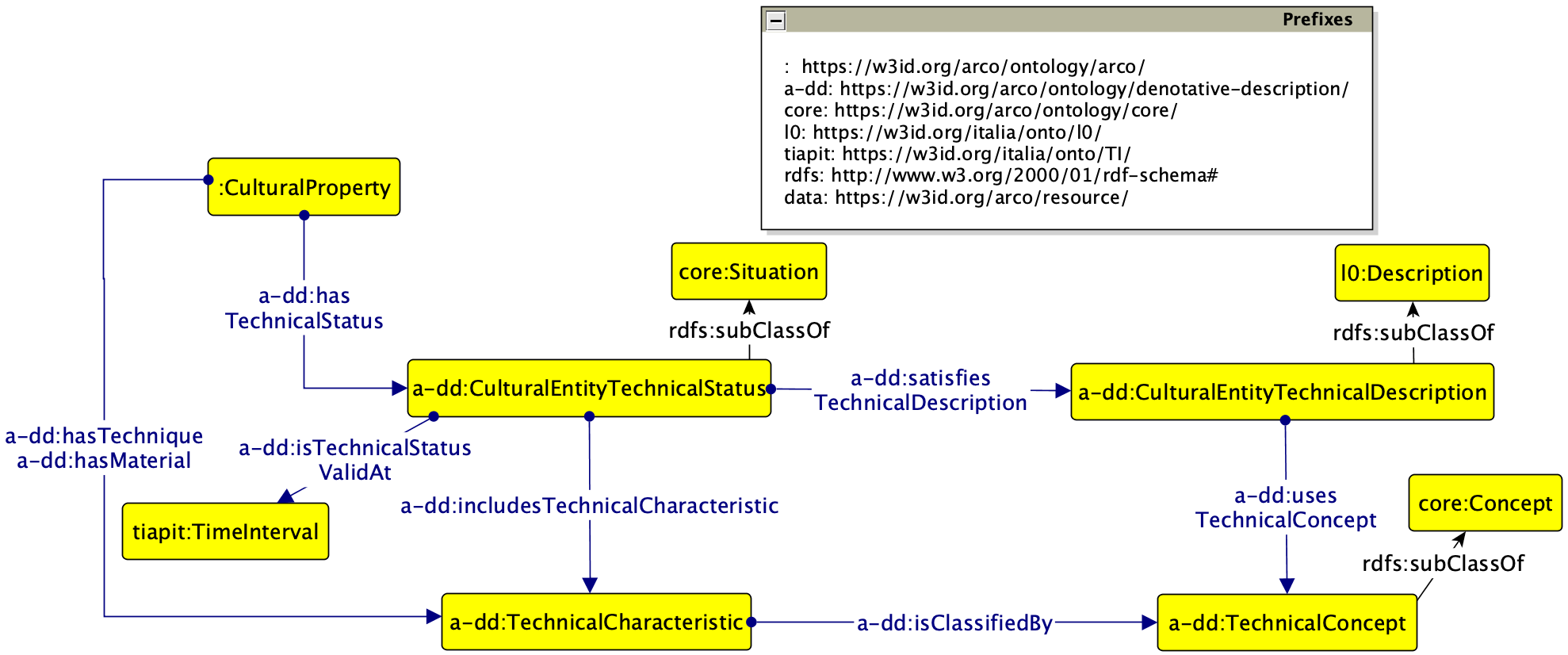}  
  \caption{The pattern DnS reused and specialised for modeling technical descriptions and status of a cultural entity.}
  \label{fig:dns-classes}
\end{subfigure}
\begin{subfigure}{\textwidth}
  \centering
  \includegraphics[width=\textwidth]{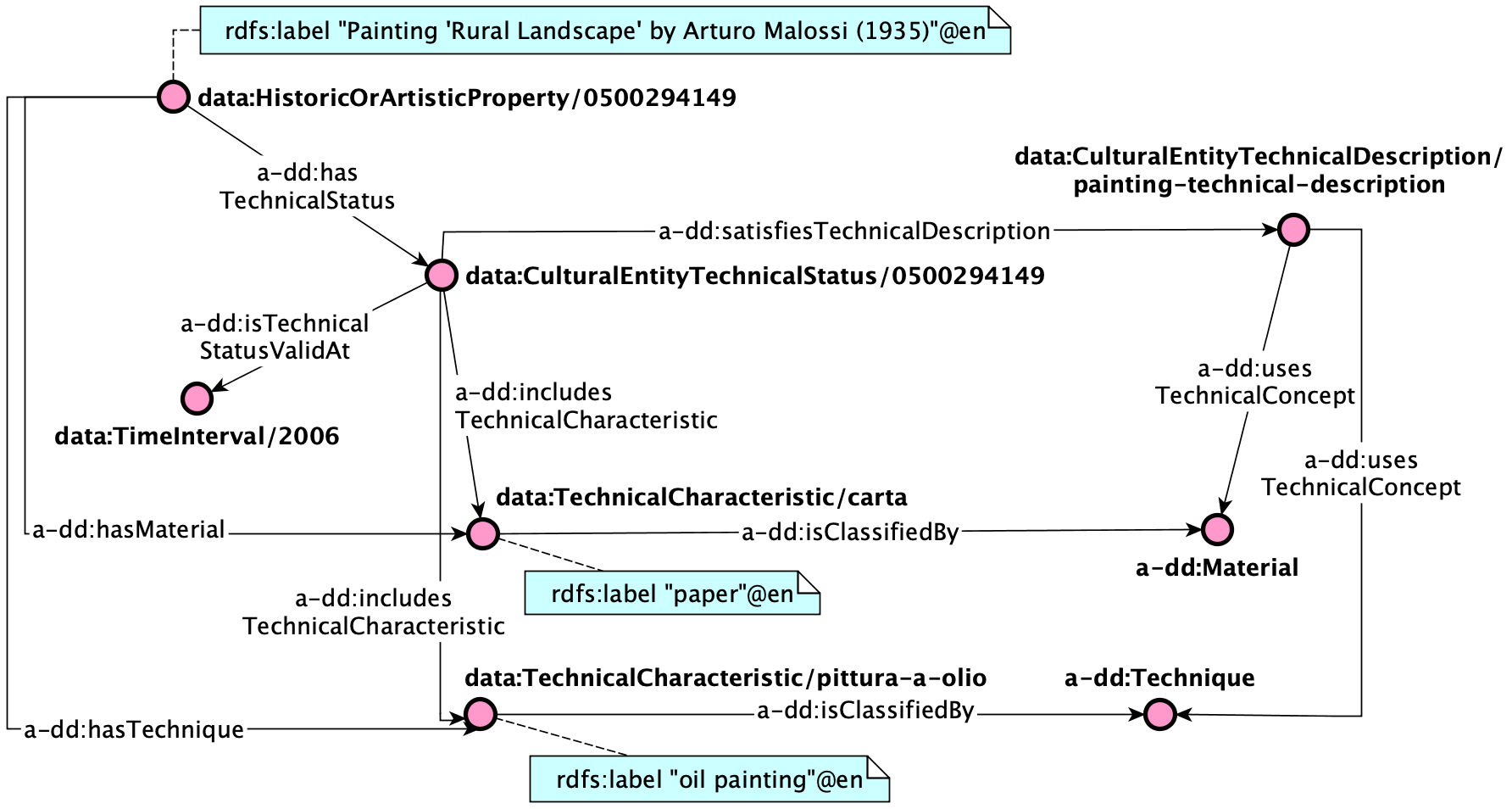}  
  \caption{An instance of the model described in Figure \ref{fig:dns-classes}, which describes the technical status of a painting made of paper (material) and realised by oil-painting (technique).}
  \label{fig:dns-instances}
\end{subfigure}
\caption{An example (in  \href{http://www.essepuntato.it/static/graffoo/specification/current.html}{Graffoo notation}) of pattern implementation in ArCo. The \emph{Situation}, \emph{Classification} and \emph{Description} ODPs are reused as templates for modeling technical characteristics of cultural properties.}
\label{img:pattern-2}
\end{figure}

An important requirement that impacted the design of ArCo consists in expressing a same concept both with  \emph{n-ary} relation patterns that enable high-level modeling needs such as temporal indexing, state changes, model evolution, meta-classes, etc., and \emph{shortcut} binary relations, which support lightweight modeling and intuitive navigation. With reference to Figures \ref{fig:dns-classes} and \ref{fig:dns-instances}, in order to represent the {\it material} of a cultural property, ArCo has: (1) a class of (reified) n-ary relations (\texttt{\href{https://w3id.org/arco/ontology/denotative-description}{a-dd:}CulturalEntityTechnicalStatus}) that include the possible \texttt{a-dd:Material}s, e.g. \texttt{data:carta} (paper), of a \texttt{:CulturalProperty}, and (2) an object property \texttt{:hasMaterial} that directly links a \texttt{:CulturalProper\-ty} to a \texttt{:Material}, and which is defined as a property chain \texttt{[a-dd:hasTechnical\-Sta\-tus O a-dd:includesTechnicalCharacteristic]} that makes it a shortcut of the n-ary relation. 

In the remainder of this Section we provide details about each of the ArCo modules, with their main concepts, the reused ODPs, and the resulting Description Logic (DL) expressivity.

\paragraph*{The \href{https://w3id.org/arco/ontology/arco}{arco} module}\label{namespace-arco} (prefix \texttt{:} and DL expressivity $\mathcal{SOIQ(D)}$) is the root of the network: it imports all the other modules. 
It formally represents top-level distinctions from the CH domain, following the definitions given in \href{http://www.iccd.beniculturali.it/it/normative}{ICCD cataloguing standards}. The top-level class
is \texttt{:CulturalProperty}, which is modeled as a 
\emph{partition} of two classes:  \texttt{:Tangible\-Cultural\-Property}, e.g. a photograph, and \texttt{:Intangible\-Cultural\-Property} e.g. a traditional dance.

\noindent\texttt{:Tangible\-Cultural\-Property} is further specialized in \texttt{:Mo\-vable\-Cul\-tu\-ral\-Pro\-per\-ty}, e.g. a painting, and \texttt{:Im\-mo\-va\-ble\-Cul\-tu\-ral\-Pro\-per\-ty}, e.g. an archaeological site. Additional, more specific types are defined down the hierarchy\footnote{Cf. the \href{https://github.com/ICCD-MiBACT/ArCo/blob/master/ArCo-release/httpd/public-html/img2/culturalproperty-classification.jpg}{diagram} on Github.}: \texttt{:DemoEthno\-Anthropologica\-lHeritage}, \texttt{:Archaeological\-Property}, \texttt{:Ar\-chi\-te\-ctural\-Or\-Land\-scape\-Her\-it\-age}, \texttt{:Historic\-OrArtistic\-Property}, \texttt{:Mu\-sic\-Her\-it\-age}, \texttt{:Natural\-Heritage}, 
\texttt{:Numismatic\-Property}, \texttt{:Photographic\-Heritage}, \texttt{:Scien\-ti\-fic\-Or\-Techno\-lo\-gi\-cal\-Her\-it\-age}, \texttt{:Historic\-OrArtistic\-Property}. \\
Other distinctions are implemented, as in between \texttt{:ComplexCultural\-Property}, e.g. a carnival costume, consisting of an aggregate of more than one \texttt{:Cultural\-PropertyComponent}, e.g. hat, trousers, etc., and  \texttt{:CulturalProperty\-Residual}, i.e. the only residual of a cultural property, such as the handle of an amphora.

\paragraph*{The \href{https://w3id.org/arco/ontology/core}{core} module} (prefix \texttt{core:} and DL expressivity $\mathcal{SHI(D)}$) represents general-purpose concepts orthogonal to the whole network, which are imported by all other ontology modules. This module reuses a number of patterns, for example the \href{http://www.ontologydesignpatterns.org/cp/owl/partof.owl}{Part-of}, the \href{http://www.ontologydesignpatterns.org/cp/owl/classification.owl}{Classification} and the \href{http://www.ontologydesignpatterns.org/cp/owl/situation.owl}{Situation} patterns.

\paragraph*{The \href{https://w3id.org/arco/ontology/catalogue}{catalogue} module} (prefix \texttt{a-cat:} and DL expressivity $\mathcal{SOIF(D)}$) provides means to represent catalogue records, and link them to the cultural properties they are a record of.
Different types of \texttt{a-cat:\-Catalogue\-Record} are defined, based on the typology of cultural property they describe. \texttt{a-cat:\-Catalogue\-Records} have \texttt{a-cat:\-CatalogueRecord\-Version}s, which are modeled by implementing the \href{http://ontologydesignpatterns.org/cp/owl/sequence.owl}{Sequence} pattern.


\paragraph*{The \href{https://w3id.org/arco/ontology/location}{location} module} (prefix \texttt{a-loc:} and DL expressivity $\mathcal{SHIF(D)}$) addresses spatial and geometrical information. A cultural property may have multiple locations, motivated by different perspectives: history, storage, finding, etc. Sometimes they coincide, sometimes they do not. Those perspectives are represented by the class \texttt{a-loc:\-Location\-Type}. A certain location type of a cultural property holds during a time interval. This concept is modeled by \texttt{a-loc:\-TimeIndex\-edTypedLocation}, which implements and specialises the \href{http://www.ontologydesignpatterns.org/cp/owl/timeindexedsituation.owl}{Time-Indexed Situation} pattern. 
This module also defines the concept of \texttt{a-loc:\-Cadastral\-Identity} of a cultural property, e.g. the cadastral unit, in which the cultural property is located.

\paragraph*{The \href{https://w3id.org/arco/ontology/denotative-description}{denotative description} module}\label{dns} (prefix \texttt{a-dd:} and DL expressivity $\mathcal{SOIQ(D)}$) encodes the characteristics 
of a cultural property, as detectable and/or detected during the cataloguing process and measurable according to a reference system. Examples include measurements e.g. length, constituting materials e.g. clay, employed techniques e.g. melting, conservation status e.g. good, decent, bad. \\
To represent those characteristics we reuse the \href{http://www.ontologydesignpatterns.org/cp/owl/descriptionandsituation.owl}{Description and Situation} and the \href{http://www.ontologydesignpatterns.org/cp/owl/classification.owl}{Classification} patterns. Figure ~\ref{img:pattern-2} shows how we model the \texttt{a-dd:CulturalEn\-tity\-TechnicalStatus}, intended as a situation in which a cultural entity (e.g. a cultural property) has some of these characteristics (e.g. is square-shaped). Each characteristic is \emph{classified} by a \texttt{a-dd:TechnicalConcept}, e.g. the \texttt{a-dd:Shape}\footnote{For the most common values we provide a controlled vocabulary.}. These concepts are used in the \texttt{a-dd:CulturalEntityTechnicalDescription}, that is the conceptualization of the relevant technical characteristics of a cultural entity (see also the beginning of Section \ref{sec:ontology} for more details).

\paragraph*{The \href{https://w3id.org/arco/ontology/context-description}{context description} module}\label{namespace-cd} (prefix \texttt{a-cd:} and DL expressivity $\mathcal{SOIQ(D)}$) represents attributes that do not result from a measurement of features in a cultural property, but are associated with it.
Examples include: information about authors, collectors, copyright holders; relations to other objects such as inventories, bibliography, protective measures, collections; activities such as surveys, conservation interventions; involvement in situations, e.g. commission, coin issuance, estimate, legal proceedings. In order to represent an \texttt{a-cd:\-Archival\-RecordSet}, i.e. fonds, series, subseries, which a cultural property is a member of, we reuse the \href{http://mklab.iti.gr/pericles/BornDigitalArchives_ODP.owl}{Born Digital Archives} pattern.

\paragraph*{The \href{https://w3id.org/arco/ontology/cultural-event}{cultural event} module} (prefix \texttt{a-ce:} and DL expressivity $\mathcal{SOIQ(D)}$) models cultural events, i.e. events involving cultural properties. It extends, with few classes and properties (e.g. \texttt{a-ce:\-Exhibition}), the \href{http://dati.beniculturali.it/cis/}{Cultural-ON} ontology.
This module provides an implementation of the \texttt{a-ce:\-RecurrentEvent} ODP, which we have defined based on ArCo's requirements\footnote{A thorough description of this new ODP is beyond the scope of this paper. It will be described in a dedicated publication, and shared on the \href{http://ontologydesignpatterns.org}{ODP portal.}}.

\subsection{ArCo dataset}
\label{sec:dataset}
\noindent ArCo dataset currently counts 169,151,644 triples. Table \ref{tab:stats} gives an overview of the dataset, indicating, for the most prominent concepts defined in ArCo, the size of the corresponding extension.

ArCo dataset provides 24,008 \texttt{owl:sameAs} axioms linking to 18,746 distinct entities in other datasets. Link discovery is limited to  authors (8,884 links) and places (9,862 links). Targets of ArCo links are: \href{https://wiki.dbpedia.org/}{DBpedia} (12,622 linked entities), \href{https://portal.dnb.de}{Deutsche National Bibliothek} (152 linked entities) \href{http://data.fondazionezeri.unibo.it/}{Zeri\&LODE} (847  linked entities), \href{http://yago-knowledge.org}{YAGO} (860 linked entities) \href{https://pro.europeana.eu/page/linked-open-data}{Europeana} (30 linked entities), \href{http://dati.isprambiente.it}{Linked ISPRA} (598 linked entities) \href{https://www.wikidata.org/wiki/Wikidata:Main_Page}{Wikidata} (2,091 linked entities), \href{https://www.geonames.org/}{Geonames} (1,466 linked entities), and from Getty vocabularies: \href{http://www.getty.edu/research/tools/vocabularies/ulan/}{ULAN} (13 linked entities) and \href{http://www.getty.edu/research/tools/vocabularies/tgn/}{TGN} (67 linked entities). 
Entity linking is performed with \href{http://aksw.org/Projects/LIMES.html}{LIMES}, configured  to use a Jaccard distance computed on the \texttt{rdfs:label} literals associated with the entities. We use an extremely selective threshold (0.9 on a [0-1] range), below which candidate links are cut off. We tested lower threshold values with manual inspection on 10\% of the produced links: 0.9 is the minimum to approximate 100\% reliability of results. The LIMES configuration files used in the linking process are available on \href{https://doi.org/10.5281/zenodo.2630565}{Zenodo}.

\begin{table}[!ht]
\centering
\caption{Dataset statistics.} 
\label{tab:stats}
\resizebox{0.87\textwidth}{!}{ 
\begin{tabular}{p{6cm}|r||p{6cm}|r}
\centering
{\bf Metric} & {\bf Result} & {\bf Metric} & {\bf Result} \\ \hline\hline
 \# instances of CulturalEntity & 822,452 & \# of triples of CulturalProperty hasAuthorshipAttribution AuthorshipAttribution & 1,428,018 \\ \hline
 
 \# instances of CulturalProperty & 781,902 & \# of instances of agents having role Author & 54,204 \\ \hline
 
 \# instances of TangibleCulturalProperty & 781,900 & \# instances of TimeIndexedTypedLocation & 1,085,521 \\ \hline
 
 \# instances of IntangibleCulturalProperty & 2 & \# instances of LocationType & 24 \\
 \hline
 
 \# instances of MovableCulturalProperty & 775,148 & avg \# of TimeIndexedTypedLocation per CulturalEntity & 1.39\\ \hline
 
 \# instances of ImmovableCulturalProperty & 6,752 & \# instances of TechnicalConcept & 22\\ \hline
 
 \# instances of HistoricOrArtisticProperty & 511,733 &  \# instances of TechnicalCharacteristic & 22,719\\ \hline
 
 \# instances of PhotographicHeritage & 20,360 &  \# of triples of CulturalEntity hasTechnicalStatus CulturalEntityTechnicalStatus & 1,084,548 \\ \hline
 
 \# instances of ArchaeologicalProperty & 149,091 & \# instances of CatalogueRecord & 781,902  \\ \hline
 
 \# instances of NaturalHeritage & 43,964 &  \# instances of CatalogueRecordsVersion & 1,767,376\\ \hline
 
 \# instances of NumismaticProperty & 17,986 & \# of triples of CulturalProperty hasCadastralIdentity CadastralIdentity & 14,683 \\ \hline
 
 \# instances of ArchitecturalOrLandscapeHeritage & 6,505 & \# of instances of CulturalEvent & 40,331 \\ \hline
 
 \# instances of ScientificOrTechnologicalHeritage & 2,687 & \# instances of Organization & 580 \\ \hline
 
 \# instances of DemoEthnoAnthropologicalHeritage & 29,576 & avg \# of Organization per CulturalProperty & 5.2 \\ 
\end{tabular}
    }
\end{table}
\section{Evaluation and Impact of ArCo}
\label{sec:impact}
\subsection{Evaluation}
\label{sec:eval}
ArCo KG has been evaluated by following the approach used in \cite{Blomqvist2009Experiments}, along the following dimensions: usability, logical consistency and requirements coverage.

\noindent{\bf Usability.} 
The numerousness of axioms and annotations, and the use of naming conventions, gives an indication of the easiness to use an ontology and understand its commitment~\cite{DBLP:conf/esws/GangemiCCL06,Blomqvist2009Experiments}. Every ArCo ontology entity has a camel-case ID, at least one label, and one comment, both in English and Italian, and is accompanied by a detailed documentation. Many classes are also annotated with examples of usage in Turtle.
The ontology contains 395 restrictions,
130 disjointness axioms,
37 alignments with 7 external ontologies. 59 classes and properties are directly reused from other ontologies.

\noindent{\bf Semantic consistency and Requirements coverage.} 
We refer to Section \ref{sec:method} for a description of the testing phase, which allows us to assess semantic consistency and requirements coverage of ArCo. We have performed: 18 tests for inference verification, which raised 3 issues; 29 tests for provoking errors, which detected 14 cases of missing axioms. 53 CQs could be converted into SPARQL queries and provide the expected results. All issues have been fixed. In addition, we received 35 issues on GitHub, solved by the design team.



%


\subsection{Potential impact of ArCo}
\label{sec:inner-impact}
ArCo first release is dated January 2018. Since then there is evidence of an emerging and growing community around it. 
A first (of a series of) webinar, attended by 10 participants, has been recently held. 
ArCo's \href{https://groups.google.com/forum/#!forum/arco-project}{mailing-list} counts 27 subscriptions and 37 threads, so far. Between beginning of January 2019 and end of March 2019, \href{https://w3id.org/arco}{ArCo release site} has been accessed 496 times by 170 distinct users. Between March 25th, 2019 and April 4th, 2019, \href{https://w3id.org/arco/sparql}{ArCo SPARQL enpoint} has been queried 1262 times and the \href{https://github.com/ICCD-MiBACT/ArCo}{GitHub page} has had 29 unique visitors and two clones.
In the last 12 months, \href{http://dati.beniculturali.it/progetto-arco-architettura-della-conoscenza/}{ArCo's official webpage} had 1084
\href{http://dati.beniculturali.it/metriche/}{unique visitors}.
We are aware of at least five organisations already using ArCo in their (independent) projects. \href{https://artsandculture.google.com/ 
}{Google Arts \& Culture}, in agreement with ICCD, is digitalising its collection of historical photographs (500.000). LOD about these pictures are modelled with ArCo and are ingested by Google Arts \& Culture from ICCD SPARQL endpoint. 
\href{https://www.regesta.com/}{Regesta.exe}\footnote{Their blog post on ArCo: \href{https://www.regesta.com/2018/03/20/data-architect-regesta-progetto-arco-architettura-della-conoscenza/}{Regesta's blog}.} uses ArCo for publishing LOD about artworks owned by \href{https://ibc.regione.emilia-romagna.it/en}{IBC-ER}.
\href{https://synapta.it/}{Synapta}\footnote{Their blog post on ArCo: \href{ https://synapta.it/blog/synapta-e-early-adopter-di-arco-architettura-della-conoscenza/}{Synapta's blog}.} 
reuses ArCo ontologies for representing musical instruments belonging to \href{http://museopaesaggiosonoro.org/sound-archives-musical-instruments-collection-samic/}{Sound Archives \& Musical Instruments Collection} (SAMIC); \href{http://ondata.it/}{OnData} works on linking data about areas of Italy hit by the earthquakes in 2016 to ArCo's data in the context of the project \href{http://ricostruzionetrasparente.it/}{Ricostruzione Trasparente}. \href{https://www.innova.puglia.it/home}{InnovaPuglia} is extending its ontologies with, and linking its \href{http://www.dati.puglia.it/lod}{LOD} to, ArCo KG. 

\section{ArCo: availability, sustainability, and licensing}
\label{sec:availability}
\noindent\textbf{Availability.} 
ArCo namespaces are introduced in Section~\ref{sec:ontology}. We create permanent URIs with the \href{https://w3id.org}{W3C Permanent Identifier Community Group}.
ArCo KG is available through  \href{http://dati.beniculturali.it/}{MiBAC's official portal} and \href{http://dati.beniculturali.it/sparql}{SPARQL endpoint}, and on \href{https://github.com/ICCD-MiBACT/ArCo}{GitHub}
(cf. Subsection~\ref{sec:howto}). 
Its ontology modules are indexed by, and can be retrieved from, \href{https://lov.linkeddata.es/dataset/lov}{Linked Open Vocabulary} (LOV).
Additionally, ArCo is published on \href{https://zenodo.org}{Zenodo}, which provides its 
DOI \href{https://doi.org/10.5281/zenodo.2630447
}{10.5281/zenodo.2630447}. ArCo's \href{https://zenodo.org/communities/arco}{community channel} on Zenodo aggregates all its material (data, experiment configurations, results, etc.). \\
\noindent\textbf{ArCo sustainability} is guardanteed by MiBAC's commitment to maintain and evolve ArCo, by following the XD methodology. In addition, CNR is committed to support and collaborate with MiBAC, based on their long-term and established collaboration as well as their shared objectives on this matter. ICCD's experts received guidelines and training for maintaining ArCo KG and for using the software for producing LOD. The docker on GitHub and its running instance online will be also maintained.
In addition to the institutional commitment, there is an active community growing around ArCo (additional information in Section \ref{sec:impact}), which contributes with both new requirements, model extensions, alignments, etc.\\
\noindent\textbf{Versioning and licensing.} ArCo is under version control on a public \href{https://github.com/ICCD-MiBACT/ArCo/tree/master/ArCo-release/ontologie}{GitHub repository}. ArCo KG license is \href{https://creativecommons.org/licenses/by-sa/4.0/}{Attribution-ShareAlike 4.0 International} (CC BY-SA 4.0).
%
\section{Cultural Heritage and Knowledge Graphs:\\ related work}
\label{sec:related}
 
Semantic technologies, in particular ontologies and LOD, are widely adopted within the CH domain for facilitating researchers, practitioners and generic users to study and consume cultural objects. Notable examples include: the \href{https://pro.europeana.eu/page/edm-documentation}{EDM}  and its \href{https://pro.europeana.eu/page/linked-open-data}{datasets}, the \href{http://www.cidoc-crm.org/}{CIDOC-CRM}, the \href{https://www.rijksmuseum.nl/en/search?ii=0&p=1}{Rijksmuseum collection}~\cite{DBLP:journals/ao/DijkshoornAOS18}, the \href{http://data.fondazionezeri.unibo.it/}{Zeri Photo Archive} \cite{DBLP:journals/jocch/DaquinoMPTV17}, the \href{http://www.getty.edu/research/tools/vocabularies/index.html}{Getty Vocabularies}, 
the \href{https://ibc.regione.emilia-romagna.it/servizi-online/lod/}{IBC-ER}, the \href{http://americanart.si.edu/collections/search/lod/about/}{Smithsonian Art Museum}, the \href{http://lodlam.net}{LODLAM}, the \href{http://openglam.org}{OpenGLAM}, the \href{https://artsandculture.google.com/}{Google Arts \& Culture}. ArCo substantially enriches the existing LOD CH cloud with invaluable data on the Italian CH and a network of ontologies addressing overlooked modelling issues. \\
Relevant related research discusses good practices for developing ontologies and LOD for CH~\cite{DBLP:journals/semweb/BoerWGOHIOS13, DBLP:conf/esws/SzekelyKYZFAG13, DBLP:conf/ekaw/AartWH10,DBLP:series/ihis/Hyvonen09,DBLP:journals/ao/DijkshoornAOS18}. ArCo draws from these lessons learnt, as well as from good practices in pattern-based ontology engineering~\cite{DBLP:books/ios/HGJKP2016}.

There are commonalities between CIDOC-CRM, EDM and ArCo. Nevertheless, CIDOC-CRM and EDM resulted insufficient against the requirements that ArCo needs to address. 
For example, modeling the diagnosis of a paleopathology in anthropological material, the coin issuance, the Hornbostel-Sachs classification of musical instruments, etc., all motivate the need for developing extended ontologies for representing CH properties. To further support this claim we compute the terminological coverage of EDM and CIDOC-CRM against ArCo CQs (cf. Section \ref{sec:method}), and compare it with ArCo's.
We model this task as an ontology matching task between an RDF vocabulary representing ArCo's CQs, and the ontologies being tested. The coverage measure is computed as the percentage of matched entities. We use Sketch Engine~\cite{Kilgarriff2004} to extract a reference vocabulary of keywords from ArCo's CQs. The result is an \href{https://doi.org/10.5281/zenodo.2633190}{RDF vocabulary of 66 terms}. 
Ontology matching is performed with Silk~\cite{Volz2009} by using the {\em substring} metric with 0.5 as threshold. The result shows the following coverage values (0: no coverage, 1: coverage): ArCo 0.68, CIDOC-CRM 0.29, EDM 0.12.
However, ArCo ontologies are aligned to (i.e. indirectly reuse) CIDOC-CRM and EDM
~\cite{DBLP:journals/semweb/IsaacH13}, as well as to \href{http://id.loc.gov/ontologies/bibframe.html}{BIBFRAME}, \href{http://vocab.org/frbr/core}{FRBR} and \href{http://purl.org/spar/fabio}{FaBiO} (for bibliographic data), and to \href{http://www.essepuntato.it/2014/03/fentry}{FEntry} and \href{http://purl.org/emmedi/oaentry}{OAEntry} (dedicated to photographs and artworks). 
Directly reused ontologies include:
\href{http://dati.beniculturali.it/cis/}{Cultural-ON} (Cultural ONtology), which models Italian 
cultural institutes, sites, and cultural events~\cite{Lodi2017}, and is maintained by MiBAC; \href{https://github.com/italia/daf-ontologie-vocabolari-controllati}{OntoPiA}, a network of ontologies and controlled vocabularies, based on \href{http://www.ontologydesignpatterns.org/ont/dul/DUL.owl}{DUL} patterns, which model top-level information crossing different domains (e.g. People, Organisation, Location) and recommended as a standard by MiBAC\footnote{OntoPia is a {\it de facto} standard for open data of the Italian Public Administration.}.

\section{Conclusion and ongoing work}
\label{sec:conclusion}
This paper presented ArCo: the knowledge graph of Italian Cultural Heritage (CH), encoded and published as linked open data (LOD). ArCo is a robust Semantic Web resource, nevertheless it is an evolving creature. As such, it can be further improved and enriched in many ways.

Concerning identity, our URI production strategy may produce possible ambiguous identifiers, i.e. same URI for different entities of the world. Because involved entities mainly represent Italian authors and organisations (cultural properties are uniquely identified, places have robust keys), we expect a small number of ambiguous cases as compared to the dimension of the dataset. Nevertheless, we are currently working on spotting them by applying a set of heuristics and validating them with the help of experts. We are also experimenting different techniques for improving external linking, including LIMES' machine learning modules, key- and linkkey-based interlinking methods~\cite{DBLP:conf/ecai/AtenciaDE14}, and crowdsourcing (involving experts) for both validation and enrichment.

%
There are aspects that are yet to be modelled: for example some specific characteristics of naturalistic heritage, e.g. slides and phials associated to an \emph{herbarium}, or the optical properties of a stone, to name a few. ArCo development includes associating pictures to each cultural entity, which are available, but not in the dataset yet.
Additional effort must be put to complete the translation of the data to other languages. At the moment, the data are expressed in Italian as from the Catalogue, 17,906,639 entities have an English label. A first step is to complete the English translation. Although this is a task to be performed by experts, we are considering supporting them with automatic translation.
In order to facilitate reuse of ArCo ontologies, we plan to develop additional tool support for CH data owners, and to address requirements coming from the library and archive domains. 

\printbibliography

\end{document}